\documentclass[12pt]{iopart}
\usepackage{iopams}
\usepackage{lineno,hyperref}
\usepackage[cmex10]{amsmath}
\usepackage{amssymb}
\usepackage[mathscr]{eucal}
\usepackage{amsbsy}
\usepackage{esint}
\usepackage{tikz}
\usepackage{graphicx}
\usepackage{multirow}
\usepackage{makecell}
\usepackage{subfigure}
\usepackage{longtable}
\usepackage{harvard}

                    {$\blacksquare$\vspace*{7pt}} 

\def\N{{\mathbb N}}

\def\f{\frac}

\def\bi{{\mathbf i}}

\def\h{{\mathbf h}}

\def\x{\boldsymbol{x}}

\def\z{{\boldsymbol z}}

\def\mD{{\mathcal D}}
\def\mE{{\mathcal E}}

\def\bi{\begin{itemize}} \def\ei{\end{itemize}}
\def\be{\begin{eqnarray*}}
\def\ee{\end{eqnarray*}}

\def\etal{{\it et al }}

\def\0{{\mathbf 0}}

\newcommand{\beq}{\begin{equation}}
\newcommand{\eeq}{\end{equation}}

\def\XXint#1#2#3{{\setbox0=\hbox{$#1{#2#3}{\int}$ }
\vcenter{\hbox{$#2#3$ }}\kern-.55\wd0}}

\modulolinenumbers[5]

\begin{document}

\title[Automated 3D cephalometric landmark identification using CT]{Automated 3D cephalometric landmark identification using computerized tomography}

\author{Hye Sun Yun\dag, Chang Min Hyun\dag \footnote[5]{To whom correspondence should be addressed
		(chammyhyun@yonsei.ac.kr)}, Seong Hyeon Baek\dag, Sang-Hwy Lee\ddag, Jin Keun Seo\dag}
\address{\dag School of Mathematics and Computing (Computational Science and Engineering), Yonsei University, Seoul, South Korea}
\address{\ddag Department of Oral and Maxillofacial Surgery, Oral Science Research Center, College of Dentistry, Yonsei University, Seoul, South Korea}

\begin{abstract}
Identification of 3D cephalometric landmarks that serve as proxy to the shape of human skull is the fundamental step in cephalometric analysis. Since manual landmarking from 3D computed tomography (CT) images is a cumbersome task even for the trained experts, automatic 3D landmark detection system is in a great need.
Recently, automatic landmarking of 2D cephalograms using deep learning (DL) has achieved great success, but 3D landmarking for more than 80 landmarks has not yet reached a satisfactory level, because of the factors hindering machine learning such as the high dimensionality of the input data and limited amount of training data due to ethical restrictions on the use of medical data.
This paper presents a semi-supervised DL method for 3D landmarking that takes advantage of anonymized landmark dataset with paired CT data being removed. The proposed method first detects a small number of easy-to-find reference landmarks, then uses them to provide a rough estimation of the entire landmarks by utilizing the low dimensional representation learned by variational autoencoder (VAE). Anonymized landmark dataset is used for training the VAE.
Finally, coarse-to-fine detection is applied to the small bounding box provided by rough estimation, using separate strategies suitable for mandible and cranium. For mandibular landmarks, patch-based 3D CNN is applied to the segmented image of the mandible (separated from the maxilla), in order to capture 3D morphological features of mandible associated with the landmarks. We detect 6 landmarks around the condyle all at once, instead of one by one, because they are closely related to each other. For cranial landmarks, we again use VAE-based latent representation for more accurate annotation. In our experiment, the proposed method achieved an averaged 3D point-to-point error of 2.91 mm for 90 landmarks only with 15 paired training data.
\end{abstract}

\maketitle
	
\section{Introduction}

Cephalometric analysis is commonly used by dentists, orthodontists, and oral and maxillofacial surgeons to provide morphometrical guidelines for diagnosis, surgical planning, growth analysis, and treatment planning by analyzing dental and skeletal relationships in the craniofacial complex \cite{Tenti1981}. It is based on cephalometric landmarks, which serve as proxy to the skull morphological data pertaining to craniofacial characteristics \cite{Proffit2018}. Conventional cephalometric analysis uses two-dimensional (2D) cephalometric radiographs (lateral and frontal radiographs), which have drawbacks including geometric distortions, superimpositions, and the dependence on correct head positioning \cite{Pittayapat2014}. Due to recent advances in image processing techniques and the need for accurate craniofacial analysis, a three-dimensional (3D) approach to the cephalometric landmarks obtaining 3D computerized tomography (CT) images is gaining preference over the conventional 2D techniques \cite{Adams2004,Nalcaci2010,Lee2014}.

Recently, there have been many studies conducted on automated cephalometric landmark identification that aim to find the landmarks and enable immediate cephalometric analysis, because manual landmarking and cephalometric analysis are labor-intensive and cumbersome tasks even for the trained experts. Due to recent advances in deep learning techniques, the automated annotation of 2D cephalometric landmarks may now be used for clinical application \cite{Arik2017,Lindner2016}. Conversely, automated 3D cephalometric tracing (for 90 landmarks) may not yet be utilized in clinical applications, wherein the required average error is commonly designated to be less than 2 mm \cite{Codari2017,Montufar2018,Lee2019,Kang2020,Yun2020}.
The high dimensionality of the input data (e.g., $512\times 512 \times 512$) and limited number of training data are the main factors that hinder the training of deep learning networks for learning the 3D landmark positional vectors from 3D CT data. Moreover, due to the current legal and ethical restrictions on medical data, it is very difficult to utilize CT data from patients.

To overcome the above-mentioned learning problems caused by the high input dimensions and training data deficiencies, the method proposed in this study utilizes semi-supervised learning that takes advantage of a large number of anonymized landmark dataset (without using the corresponding CT dataset) which have been used in surgical planning and treatment evaluation. We use these landmark dataset to obtain their low dimensional representations, reducing the dimensions of the total landmark vectors ($270=90\times 3$ dimension) to only 9 latent variables via a variational autoencoder (VAE) \cite{Kingma2013}. For training the VAE, a normalized landmark dataset is used to efficiently learn skull shape variations while ignoring unnecessary scaling factors. With this dimensionality reduction technique, the positions of all 90 landmarks can be roughly estimated by identifying a small number of easy-to-find reference landmarks (10 landmarks), which can be accurately and reliably identified via a simple deep learning method \cite{Lee2019}.

The rough estimation of all landmarks is used to provide a small 3D bounding box for each landmark in the 3D CT images. Following this, we apply convolutional neural networks (CNNs) to these small bounding boxes to enable the accurate placement of landmarks. Our fine detection strategy is divided into two parts: mandible and cranium. It is desirable to accurately capture the morphological variability of the mandible because the shape of the mandible can be affected by a variety of factors, including the masticatory occlusal force, muscular force, functional activity such as breathing and swallowing, and age \cite{Vallabh2019}. Noting that landmarks on the mandible represent morphological features of a 3D mandibular surface geometry, we apply 3D CNN to a segmented image of the mandible (separated from the cranium). We refer to the recent study \cite{Jang2020} for a segmentation method to separate the mandible from the cranium.

Because several landmarks around the condyle are closely related to each other, it is better to detect these landmarks all at once. For the landmarks on the midsagittal plane, it is better to further reduce the dimensionality of the input by using a partially integrated 2D image of the midsagittal plane. For the remaining landmarks lying on the cranium, we again use the anonymized landmark dataset to obtain a more accurate latent representation of all landmarks on the cranium, due to its rigidity. The proposed approach achieved an average 3D point-to-point error of 2.91 mm for 90 landmarks, which nearly meets the clinically acceptable precision standard. It should be emphasized that this accuracy has been achieved using a very small amount of training data.

\section{Method}
For ease of explanation, we begin by introducing the following notations. Five easy-to-find reference landmarks (CFM, Bregma, Na, and Po (L/R)) are used as  the basis for constructing a coordinate system to determine the midsagittal and axial planes, and they were utilized for data normalization (methods for obtaining these five reference landmarks will be described in Section \ref{stage1}).
\begin{itemize}
	\item $\x$ denotes a 3D CT image, which is defined on a voxel grid $\Omega:=\{v =(v_1,v_2,v_3): v_j=1,\cdots, 512 \mbox{ for } j = 1,2,3 \}$. Here, we set $v_1$ as the normal direction of the midsagittal plane.
	\item $\x_b$ denotes a binarized CT image of $\x$, defined by
	\begin{equation} \label{binarizing}
		\x_b = \left\{\begin{array}{cl} \x_b(v_1,v_2,v_3) = 1 & \mbox{ if } \x(v_1,v_2,v_3) \geq \rho  \\  \x_b(v_1,v_2,v_3) = 0 & \mbox{ otherwise } \end{array} \right.
	\end{equation}
	where $\rho$ is a thresholding value.
	\item $\x^{\mbox{\tiny mid}}$ denotes a partially integrated 2D image of $\x_b$ in the normal direction of the midsagittal plane, defined by
	\begin{equation} \label{partialsum}
		\x^{\mbox{\tiny mid}} = \sum_{v_1=a}^{b} \x_b(v_1,v_2,v_3)
	\end{equation}
	where $[a,b]$ determines the truncated volume of $\x_b$.
	\item $\mathfrak{R}^{\mbox{\tiny cr}} \in \mathbb{R}^{138(=46\times 3)}$ and $\mathfrak{R}^{\mbox{\tiny md}}  \in \mathbb{R}^{132(=44\times 3)}$ denote the concatenated vectors of 46 cranial and 44 mandibular 3D landmarks, respectively. The entirety of the landmarks $\mathfrak{R}\in\mathbb R^{270(=90\times3)}$ is defined by $\mathfrak{R}:=[\mathfrak{R}^{\mbox{\tiny cr}}, \mathfrak{R}^{\mbox{\tiny md}}]$. See \ref{appenA} for more detailed information of the landmarks.
	\item $\mathfrak{R}^{\mbox{\tiny cr}}_{\sharp}\in \mathbb{R}^{24(=8\times 3)}$ denotes a concatenated vector of landmarks (Bregma, CFM, Na, ANS, Or (L/R), and Po (L/R)) in the cranium and $\mathfrak{R}^{\mbox{\tiny md}}_{\sharp}\in \mathbb{R}^{6(=2\times 3)}$ denotes a concatenated vector of landmarks (MF (L/R)) in the mandible. A reference landmark vector $\mathfrak{R}_{\sharp}\in \mathbb R^{30(=10\times3)}$ is defined by $\mathfrak{R}_{\sharp}=[\mathfrak{R}^{\mbox{\tiny cr}}_{\sharp}, \mathfrak{R}^{\mbox{\tiny md}}_{\sharp}]$.
\end{itemize}

\begin{figure}[t!]
	\centering
	\includegraphics[width=1\textwidth]{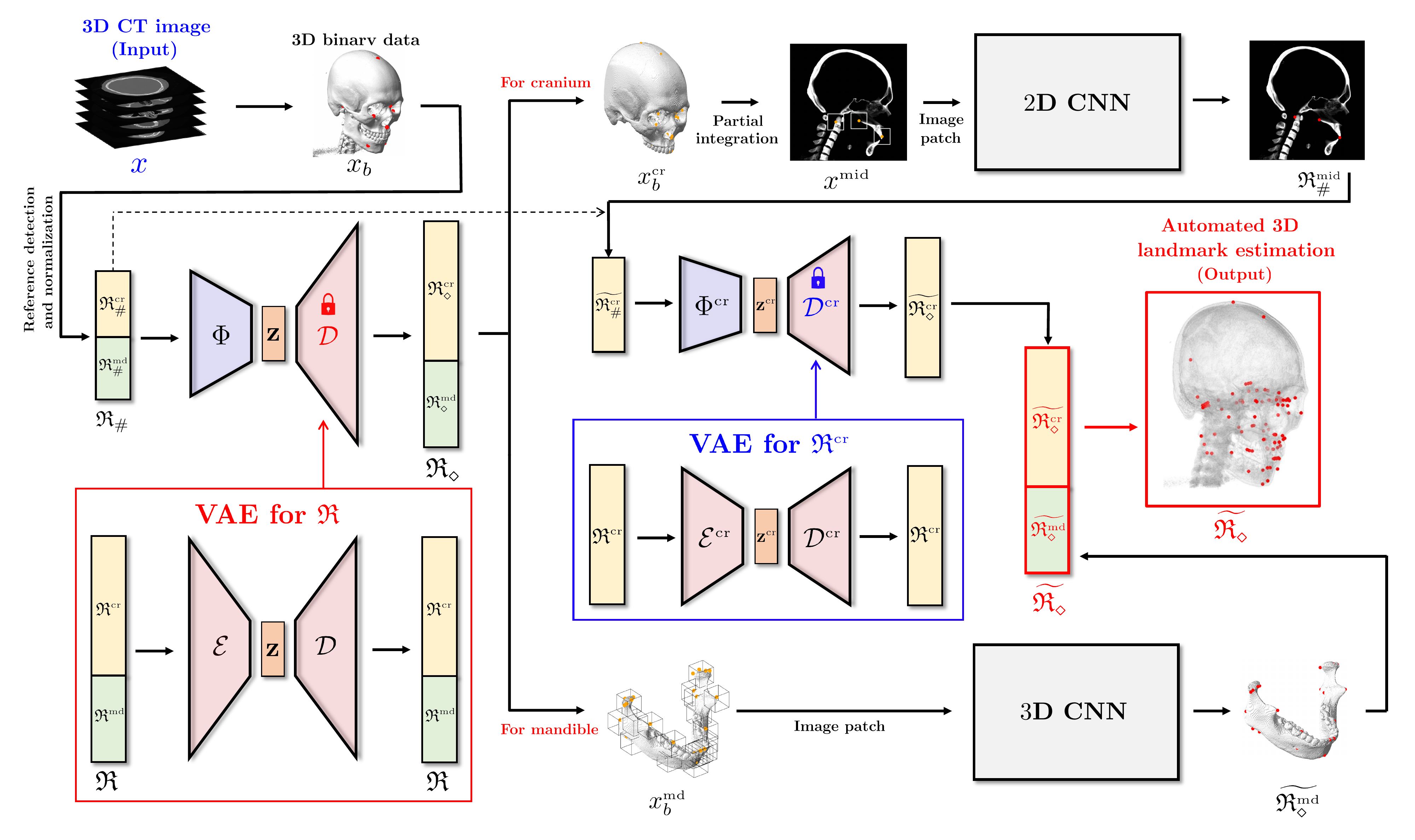}
	\caption{Schematic diagram of the proposed method for the 3D landmark annotation system. }
	\label{whole}
\end{figure}

The 3D cephalometric landmarking aims to develop a function $f:\x \mapsto \mathfrak{R}$ that maps from a 3D CT image $\x$ to all landmarks $\mathfrak{R}$.
To learn the landmark detection map $f$, deep learning techniques can be used. Unfortunately, due to legal and ethical restrictions on medical data, a few paired data are available. This severe shortage of paired data makes it difficult to obtain an accurate and reliable map $f:\x \mapsto \mathfrak{R}$ in the following supervised learning framework:
\begin{align}\label{DL-3D}
	f=\underset{f\in \N \mbox{\scriptsize et}}{\mbox{argmin}}\f{1}{N_p}\sum_{i=1}^{N_p}\|f (\x^{(i)})- \mathfrak{R}^{(i)}\|^2_2,
\end{align}
where $N_p$ is the small number of paired training data, $\{(\x^{(i)}, \mathfrak{R}^{(i)}) : i=1,\cdots,N_p\}$ is a paired dataset, $\N\mbox{et}$ is a deep learning network, and $\|\cdot\|$ is the standard Euclidean norm. In our study, only 15 paired data are available (i.e., $N_p=15$). Even with a certain amount of paired data, the learning process \eqref{DL-3D} of the direct detection map $f$ can be difficult because the dimension of the input image is very large (greater than 100,000,000).

The proposed method attempts to address this problem by taking advantage of a semi-supervised learning framework that permits the utilization of the $N_{l}$ number of anonymized landmark data $\{\mathfrak{R}^{(N_p+i)}\}_{i=1}^{N_{l}}$ whose corresponding CT data are not provided. As shown in Figure \ref{whole}, the proposed method comprises the following three main steps: (i) To obtain easy-to-find reference landmarks $\mathfrak{R}_{\sharp}$, we apply CNN with 2D illuminated images generated from a binarized CT image $\x_b$ and normalize the output with respect to the cranial volume. (ii) A rough estimation of entire landmarks $\mathfrak R$  is obtained using the partial knowledge $\mathfrak{R}_{\sharp}$ and a VAE-based low dimensional representation of $\mathfrak R$. (iii) Using this estimation, coarse-to-fine detection for $\mathfrak R$ is conducted, wherein separate strategies are utilized for the mandibular and cranial landmarks. For the mandibular landmarks, the landmarks are accurately identified by applying 3D patch-based CNNs to capture the morphological features on a 3D surface geometry associated with the landmarks, wherein an input patch is extracted based on the coarse estimation. For cranial landmarks, we first detect three landmarks lying on the midsagittal plane by applying a 2D CNN whose input is an extracted 2D patch from a partially integrated image $\x^{\mbox{\tiny mid}}$ in basis of the coarse estimation. By utilizing the three finely-detected landmarks and cranial reference landmarks $\mathfrak R^{\mbox{\tiny cr}}_{\sharp}$ as the partial information of $\mathfrak R^{\mbox{\tiny cr}}$, the remaining cranial landmarks are finely annotated via a VAE-based local-to-global estimation utilizing the same method in the previous step .

Each of these steps is described in detail as follows.

\subsection{Detection of easy-to-find reference landmarks and uniform scaling for skull normalization with respect to the cranial volume} \label{stage1}

\begin{figure}[t!]
	\centering
	\includegraphics[width=.9\textwidth]{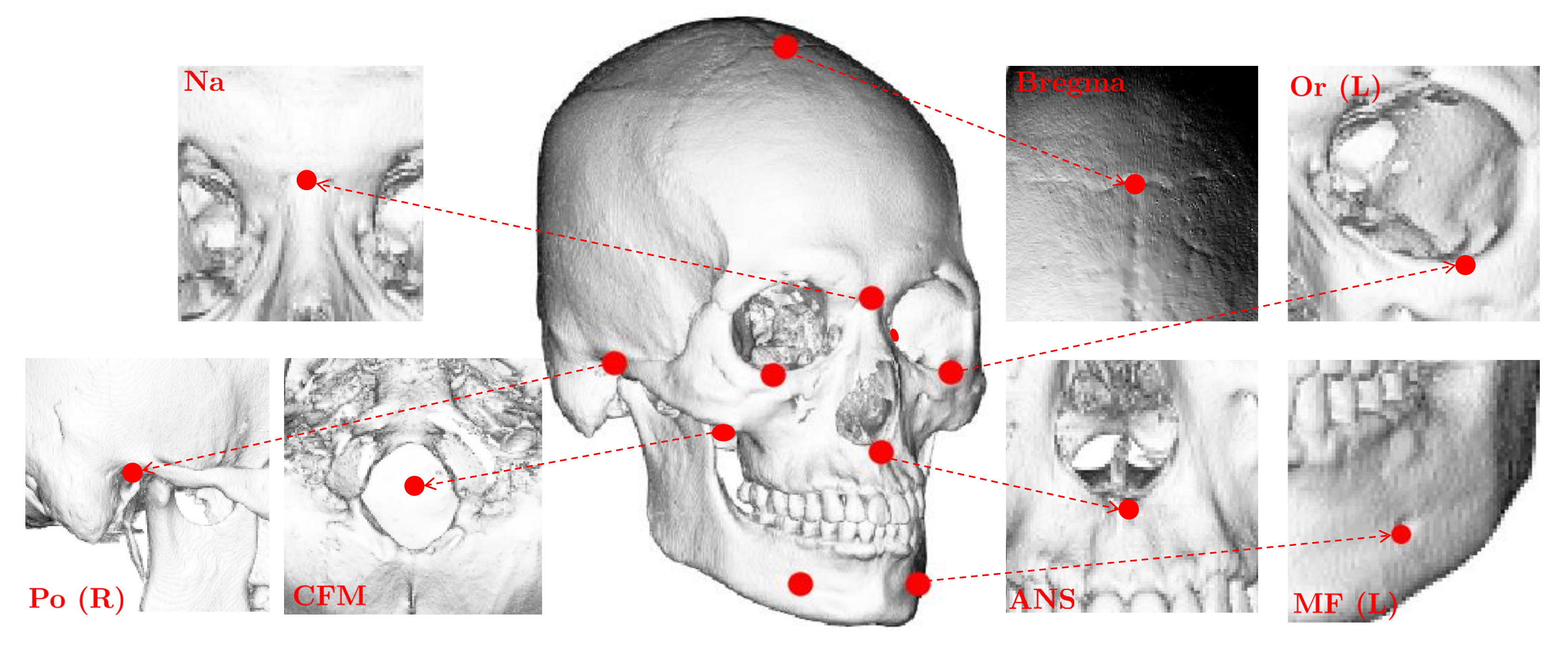}
	\caption{Reference landmarks. These are easy-to-find through CNN with input of the illuminated images because they have strong geometric cues that can be revealed in illuminated 2D images.}
	\label{shadowed_img_reference}
\end{figure}

The first step of the proposed method is to find 10 reference landmarks $\mathfrak R_{\sharp}$ from a given $\x$. Initially, a CT image $\x$ is converted into a binarized image $\x_b$ by \eqref{binarizing}. From $\x_b$, 2D illuminated images are generated by manipulating various lighting and viewing directions (see Figure \ref{shadowed_img_reference}). By applying VGGNet \cite{Simonyan2014} to these illuminated images, the reference landmarks $\mathfrak R_{\sharp}$ are accurately and automatically identified. This detection method is based on that presented in the recent study \cite{Lee2019}.

Using these identified reference landmarks, data normalization is conducted for efficient feature learning of skull shape variations in further steps. By applying uniform scaling with respect to the cranial volume, the landmark vector $\mathfrak R_{\sharp}$ is normalized, wherein the cranial volume is defined via a product of the distance between the $v_1$-coordinate of Po (L) and Po (R) (cranial length), the distance between the $v_2$-coordinate of Po (L) and Na (depth), and the distance between the $v_3$-coordinate of CFM and Bregma (height). This data normalization minimizes the positional dependencies of landmarks on the translation, rotation, and overall size of the skull; therefore, shape information of the skull (regarding facial deformities) can be effectively learned in further VAE-based steps. From here on, we will denote all landmark vectors as normalized vectors (e.g., $\mathfrak{R}$ and $\mathfrak{R}_{\sharp}$ are normalized vectors for total landmarks and reference landmarks).

\subsection{Rough estimation of all landmarks from reference landmarks using VAE-based low dimensional representation} \label{stage2}
\begin{figure}[t!]
	\centering
	\includegraphics[width=1\textwidth]{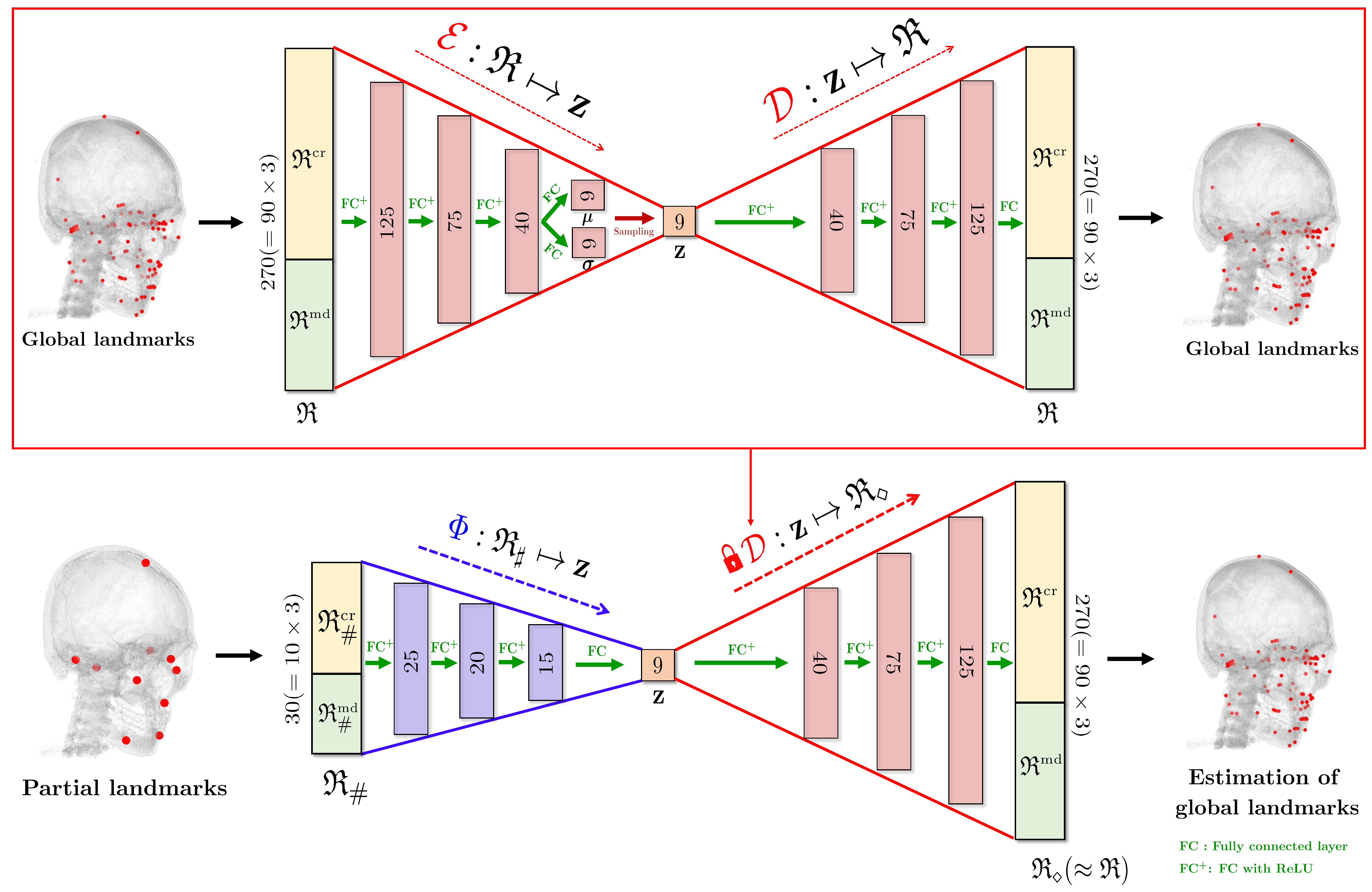}
	\caption{Initial estimation of all 90 landmarks $\mathfrak R$ using the knowledge of 10 reference landmarks $\mathfrak R_{\sharp}$. This is possible because all landmarks $\mathfrak R$ can be roughly represented by only 9 latent variables.}
	\label{pL_to_fL}	
\end{figure}

This section provides a method for roughly estimating all landmarks $\mathfrak R$ from the reference landmarks $\mathfrak R_{\sharp}$ that are accurately annotated in the previous step. Based on the method in \cite{Yun2020}, we build a bridge that connects $\mathfrak R_{\sharp}$ and $\mathfrak R$ by taking advantage of a low dimensional representation of $\mathfrak R$ learned by a variational autoencoder (VAE) \cite{Kingma2013}.

The rough estimation obtained from $\mathfrak R_{\sharp}$, denoted by $\mathfrak R_\diamond$, is given by
\begin{equation}
	\mathfrak R_\diamond = \mathcal D \circ \Phi(\mathfrak R_{\sharp})
\end{equation}
where $\mathcal D \circ \Phi : \mathfrak R_{\sharp} \mapsto \mathfrak R_\diamond$ is a local-to-global landmark estimation map as described in Figure \ref{pL_to_fL}. The map $\mathcal D \circ \Phi$ is constructed via the following process.

Using the landmark dataset $\{\mathfrak R^{(i)}\}_{i=1}^{N_t}$, we aim to represent all landmarks $\mathfrak R$ in terms of $d$-dimensional latent variables $\mathbf{z}\in \mathbb{R}^{d}$ (with $d<<270$) by learning an encoder $\mE: \mathfrak{R} \mapsto \mathbf{z}$ and a decoder $\mD: \mathbf{z} \mapsto \mathfrak{R}$ via the following energy minimization equation:
\begin{equation}\label{VAE_mini}
	\small (\mE, \mD)=  \underset{(\mE,\mD) \in \mathbb{VAE}}{\mbox{argmin}}\sum_{i=1}^{N_t} (\| \mD\circ\mE (\mathfrak{R}^{(i)}) -\mathfrak{R}^{(i)}\|^2_2
	+D_{KL}(\mathcal{N}(\mu^{(i)},\Sigma^{(i)} ) \parallel \mathcal{N}(0,I)))
\end{equation}
where $N_t= N_p+N_{l}$ is the total number of training landmark data, $\mathbb{VAE}$ is a class of functions in the form of a given VAE network, $\mathcal{N}(\mu^{(i)},\Sigma^{(i)} )$ is a $d$-dimensional normal distribution with a mean $\mu^{(i)}$ and a diagonal covariance matrix $\Sigma^{(i)}=\mbox{diag}((\sigma^{(i)}(1))^2, \cdots, (\sigma^{(i)}(d))^2)$, $\mathcal{N}(0,I)$ is a standard normal distribution, and the last term in the loss function is the Kullback-Leibler (KL) divergence defined by:
\begin{equation} \label{KL2}
	D_{KL}(\mathcal{N}(\mu^{(i)},\Sigma^{(i)}  ) \parallel \mathcal{N}(0,I)) = \frac{1}{2} \sum _{l=1}^{d} (\mu^{(i)}(l)^2+ \sigma^{(i)}(l)^2 - \log \sigma^{(i)}(l) -1)
\end{equation}
Here, $\mu^{(i)} = (\mu^{(i)}(1), \cdots, \mu^{(i)}(d))$ and $\sigma^{(i)} = (\sigma^{(i)}(1),\cdots, \sigma^{(i)}(d))$ are the mean and standard deviation vectors obtained in the interim of the encoding process of an $i$-th training data $\mathfrak{R}^{(i)}$ (i.e., $\mE(\mathfrak{R}^{(i)})$).

The encoder $\mE$ can be expressed in the following nondeterministic form:
\begin{equation}
	\mE(\mathfrak{R}) = \textbf{z} :=\mu +\sigma\odot\h_{\mbox{\tiny noise}}
\end{equation}
where $\h_{\mbox{\tiny noise}}$ is a noise sampled from $\mathcal{N}(0,I)$, $\odot$ is the Hadamard product (i.e., element-wise product), and vectors $\mu$ and $\sigma$ are given by:
\begin{equation}
	\mu = \textbf{E}_4^\mu \textbf{h}, \sigma = \textbf{E}_4^\sigma \textbf{h}, \textbf{h} = \mbox{ReLU}(\textbf{E}_3\mbox{ReLU}(\textbf{E}_2\mbox{ReLU}(\textbf{E}_1\mathfrak R)))
\end{equation}
Here, the matrices $\{\textbf{E}_1, \textbf{E}_2, \textbf{E}_3, \textbf{E}_4^\mu, \textbf{E}_4^\sigma \}$ represent fully-connected layers and ReLU is an element-wise activation function defined by $\mbox{ReLU}(t)=\mbox{max}(t,0)$. The decoder $\mD$ is the reverse process of the encoder $\mE$, which can be represented by:
\begin{equation}
	\mD(\textbf{z}) = \textbf{D}_1\mbox{ReLU}(\textbf{D}_2\mbox{ReLU}(\textbf{D}_3\mbox{ReLU}(\textbf{D}_4 \textbf{z})))
\end{equation}
where the matrices $\{\textbf{D}_1, \textbf{D}_2, \textbf{D}_3, \textbf{D}_4\}$ represent fully-connected layers. The detailed network architecture is described in Figure \ref{pL_to_fL}.

After training the functions $\mE$ and $\mD$, a nonlinear map $\Phi:\mathfrak{R}_\sharp \to \textbf{z}$ is learned, which connects reference landmarks $\mathfrak R_\sharp$ with a latent variable $\textbf{z}=\mE(\mathfrak R)$. It is achieved by the following energy minimization equation:
\begin{equation}\label{nonli_mini}
	\Phi =  \underset{\Phi \in \mathbb{FC}}{\mbox{argmin}} \sum_{i=1}^{N_t} \| \Phi(\mathfrak{R}_{\sharp}^{(i)})-\z^{(i)} \|^2_{2}
\end{equation}
where $\z^{(i)} = \mE(\mathfrak R_{\sharp}^{(i)})$ can be obtained by the trained encoder $\mE$ and $\mathbb{FC}$ is a set of all functions that can be learned via the fully-connected network structure. The detailed architecture of the map $\Phi$ is described in Figure \ref{pL_to_fL}.

The training of $\mD\circ\Phi$ uses only the landmark dataset $\{ \mathfrak R^{(i)}\}_{i=1}^{N_t}$. Based on patterns learned from the landmark dataset, we are able to estimate all landmarks $\mathfrak R$ from the partial knowledge $\mathfrak R_{\sharp}$. By utilizing the given CT image data, therefore, we put a final touch on the rough estimation $\mathfrak R_\diamond$ to achieve more accurate landmark placement.

\subsection{Coarse-to-fine detection}
This subsection explains coarse-to-fine detection obtained using the initial estimation $\mathfrak R_{\diamond}$, which is obtained in the previous step. The coarse-to-fine detection is based on suitable strategies that rely on locations of landmarks (i.e. on the mandible or cranium).

In the binarized image $\x_b$, we separate the mandible from cranium. Kindly refer to the paper \cite{Jang2020} for the segmentation method. Let $\x_{b}^{\mbox{\tiny cr}}$ and $\x_{b}^{\mbox{\tiny md}}$ denote the separated cranial and mandibular images (as shown in Figure \ref{whole}). Using these images and the rough estimation $\mathfrak R_{\diamond}$, the following fine detection processes are conducted.

\subsubsection{Detection of mandibular landmarks}\label{md_fine_detec}

\begin{figure}[t!]
	\centering
	\includegraphics[width=1\textwidth]{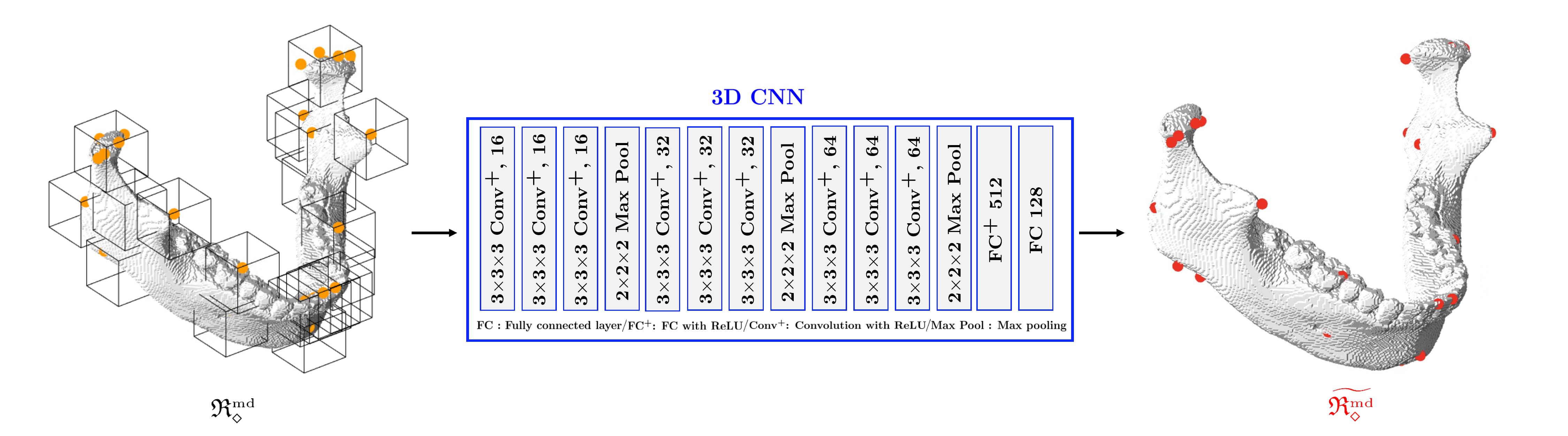}
	\caption{Mandibular landmarks detection. Patch-based 3D CNN is applied to the segmented image of the mandible (separated from the maxilla), in order to capture 3D morphological features of mandible associated with the landmarks. For six landmarks on condyle, we detect them all at once, instead of one by one, because they are positionally related to each other.}
	\label{coarse_to_fine_md}
\end{figure}

For landmarks on the mandible being articulated to the skull, a patch-based 3D CNN is applied to capture the morphological variability of the 3D mandibular surface geometry associated with the landmarks. Let $\mathfrak{R}_\diamond^j \in \mathbb{R}^3$ be a roughly estimated position of a landmark with index $j$ in $\mathfrak{R}_\diamond$. See \ref{appenA} for details of the landmark index. For each mandibular landmark (i.e., $j \in \{ 49,\cdots,90 \}$), we extract a 3D image patch $(\x_b^{\mbox{\tiny md}})_{(\eta,\mathfrak{R}_\diamond^j)}$, which is defined by a cube whose edge length is $\eta$ and center is $\mathfrak{R}_\diamond^j$. By using 3D CNN, we obtain a map $f^{\mbox{\tiny md}}_j : (\x_b^{\mbox{\tiny md}})_{(\eta,\mathfrak{R}_\diamond^j)} \mapsto \widetilde{\mathfrak{R}_{\diamond}}^j$, where $\widetilde{\mathfrak{R}_{\diamond}}^j$ is an accurate positional estimation for the landmark with index $j$ (i.e., $\widetilde{\mathfrak{R}_{\diamond}}^j \approx \mathfrak{R}^{j}$).

To learn the fine detection map $f^{\mbox{\tiny md}}_j$, we generate a training dataset by using the paired dataset $\{((\x_b^{\mbox{\tiny md}})^{(i)}, \mathfrak{R}^{(i)})\}_{i=1}^{N_p}$ as follows. For a given index $j$, 3D patches with edge lengths of $\eta$ are extracted from $(\x_b^{\mbox{\tiny md}})^{(i)}$ by varying the center location of patch in basis of label landmark position $(\mathfrak R^{(i)})^j$. As a result, the following dataset is obtained.
\begin{equation} \label{patch_training_3D}
	\{ ((\x_b^{\mbox{\tiny md}})_{(\eta,(\mathfrak{R}^{(i)})^{j}+\textbf k)}^{(i)}, (\mathfrak R^{(i)})^j ) : k_1,k_2,k_3 = -\gamma,\cdots,\gamma \mbox{ and } i=1,\cdots,N_p\}
\end{equation}
where $\textbf k \in \mathbb{R}^3 = (k_1,k_2,k_3)$ and $\eta$ is the maximum length of the center position variation. Using this dataset, the 3D CNN is trained as follows:
\begin{equation}
	f^{\mbox{\tiny md}}_j = \underset{f^{\mbox{\tiny md}}_j \in \mathbb{C}\mathbb{N}\mathbb{N}}{\mbox{argmin}} ~~ \sum_{k_1,k_2,k_3 = -\gamma}^{\gamma} \sum_{i=1}^{N_p} ~ \| f((\x_b^{\mbox{\tiny md}})_{(\eta,(\mathfrak{R}^{(i)})^{j}+\textbf k)}^{(i)}) - (\mathfrak{R}^{(i)})^{j} \|_{2}^2
\end{equation}
where $\mathbb{C}\mathbb{N}\mathbb{N}$ is a class of functions in the form of a 3D CNN.  As seen in Figure \ref{coarse_to_fine_md}, the 3D CNN is used that possesses an architecture modified from VGGNet \cite{Simonyan2014}.

In practice, several landmarks are identified in a group at once. We simultaneously identify six landmarks on the condyle (COR, MCP, LCP, Cp, Ct-in, and Ct-out), which are positionally related to one another, as well as landmarks with bilaterality (e.g. left/right mandibular foramen), which are associated with the symmetric structure of the mandible. For this group detection, we construct a 3D CNN to produce a concatenated vector of all landmark positions on the same group from one 3D image patch.

\subsubsection{Detection of cranial landmarks} \label{cr_fine_detec}

\begin{figure}[t!]
	\centering
	\includegraphics[width=1\textwidth]{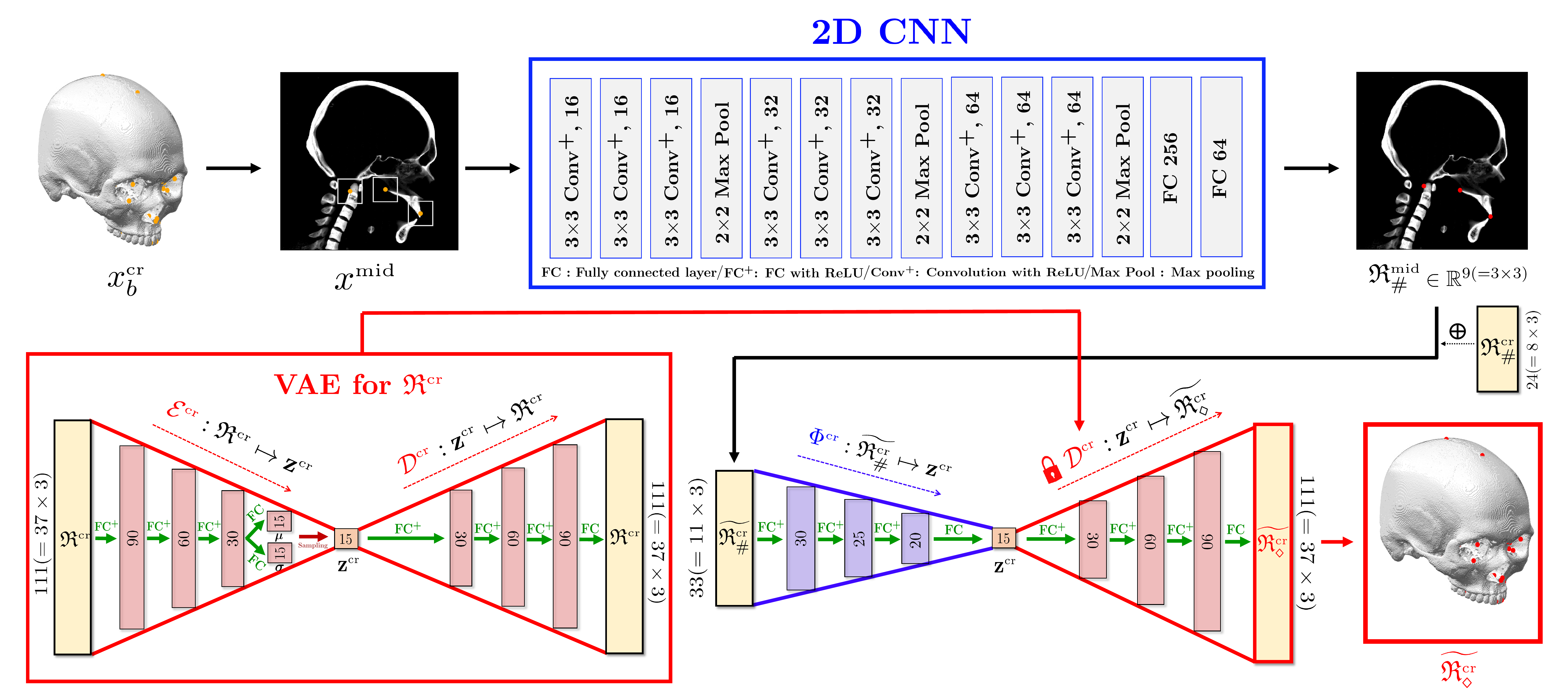}
	\caption{3D cranial landmark detection using VAE-based low dimensional representation combined with easy-to-find landmarks. Here, the entire cranial landmarks $\mathfrak R^{\mbox{\tiny cr}}$ are estimated directly from the knowledge of the reference landmarks $\mathfrak R^{\mbox{\tiny cr}}_{\sharp}$ and three landmarks $\mathfrak R_{\sharp}^{\mbox{\tiny mid}}$ on midsagittal plane that are obtained by 2D CNN.}
	\label{coarse_to_fine_cr}
\end{figure}

Landmarks on the cranium that demonstrates rigidity have less variability between subjects. According to \cite{Yun2020}, cranial landmarks have smaller variance compared to mandibular landmarks with the normalization presented in Section \ref{stage1}. Moreover, our empirical experiment shown in Figure \ref{vae_graph} demonstrates that the rough local-to-global estimation achieved using the VAE-based low dimensional representation provides more accurate annotations for cranial landmarks. Therefore, we again utilize a VAE-based low dimensional representation in the same manner as in Section \ref{stage2} by using only the cranial landmarks $\mathfrak R^{\mbox{\tiny cr}}$. To increase the detection accuracy, we enrich the partial knowledge of $\mathfrak R^{\mbox{\tiny cr}}$ by accurately detecting three additional cranial landmarks lying near the midsagittal plane (MxDML, Od, and PNS) based on the rough estimation $\mathfrak R_{\diamond}$. The overall process is illustrated in Figure \ref{coarse_to_fine_cr}.

First, we compute a partially integrated image $\x^{\mbox{\tiny mid}}$ from $\x_{b}^{\mbox{\tiny cr}}$ using \eqref{partialsum} so that the center of the truncated volume of $\x_{b}^{\mbox{\tiny cr}}$ lies on the midsagittal plane. Next, a 2D patch $(\x^{\mbox{\tiny mid}})_{(\eta,\mathfrak{R}_\diamond^j|_{v_2,v_3})}$ is extracted, which is defined by a square whose edge length is $\eta$ and center is given by $\mathfrak{R}_\diamond^j|_{v_2,v_3}$. Here, $\mathfrak{R}_\diamond^j|_{v_2,v_3}$ is a vector eliminating the $v_1$ component in the $\mathfrak{R}_\diamond^j$ and $j \in \{24,25,26\}$. Using a 2D CNN, we learn a function $f^{\mbox{\tiny cr}}_j$, which infers an accurate position of a landmark $\mathfrak{R}^j$ in $v_2$- and $v_3$-coordinates ($\mathfrak{R}^j|_{v_2,v_3}$) from the 2D image patch $(\x^{\mbox{\tiny mid}})_{(\eta,\mathfrak{R}_\diamond^j|_{v_2,v_3})}$. The landmark position in the $v_1$-coordinate is determined by the location of the midsagittal plane.

In the similar manner as in \eqref{patch_training_3D}, the following training dataset is generated.
\begin{equation} \label{patch_training_2D}
	\small{\{ ((\x^{\mbox{\tiny mid}})_{(\eta,(\mathfrak{R}^{(i)})^{j}|_{v_2,v_3}+\textbf k)}^{(i)}, (\mathfrak R^{(i)})^j|_{v_2,v_3} ) : k_1,k_2 = -\gamma,\cdots,\gamma \mbox{ and } i=1,\cdot,N_p\}}
\end{equation}
where $\textbf k \in \mathbb{R}^2 = (k_1,k_2)$ and $\eta$ is the maximum length of the center position variation. With the training dataset, the 2D CNN is trained as follows:
\begin{equation}
	f^{\mbox{\tiny cr}}_j = \underset{f^{\mbox{\tiny cr}}_j \in \mathbb{C}\mathbb{N}\mathbb{N}}{\mbox{argmin}} \sum_{k_1,k_2 = -\gamma}^{\gamma} \sum_{i=1}^{N_p} \| f((\x^{\mbox{\tiny mid}})_{(\eta,(\mathfrak{R}^{(i)})^{j}|_{v_2,v_3}+\textbf k)}^{(i)}) - (\mathfrak{R}^{(i)})^{j}|_{v_2,v_3} \|_{2}^2
\end{equation}
where $\mathbb{C}\mathbb{N}\mathbb{N}$ is a class of functions in the form of a 2D CNN. The architecture of the 2D CNN is modified from VGGNet \cite{Simonyan2014}, as illustrated in Figure \ref{coarse_to_fine_md}.

Let $\mathfrak R_{\sharp}^{\mbox{\tiny mid}}$ be a concatenated positional vector with cranial reference landmarks $\mathfrak R^{\mbox{\tiny cr}}_{\sharp}$ and three finely detected landmarks obtained by $f^{\mbox{\tiny cr}}_j$. Using this partial knowledge $\mathfrak R_{\sharp}^{\mbox{\tiny mid}}$, we find accurate cranial landmark positions $\widetilde{\mathfrak R^{\mbox{\tiny cr}}_{\diamond}}$ via
\begin{equation}
	\widetilde{\mathfrak R^{\mbox{\tiny cr}}_{\diamond}} = \mD^{\mbox{\tiny cr}} \circ \Phi^{\mbox{\tiny cr}}(\mathfrak R_{\sharp}^{\mbox{\tiny mid}})
\end{equation}
where  $\Phi^{\mbox{\tiny cr}} : \mathfrak R_{\sharp}^{\mbox{\tiny mid}} \mapsto \textbf{z}^{\mbox{\tiny cr}}$ is a nonlinear map and $\mD^{\mbox{\tiny cr}} : \textbf{z}^{\mbox{\tiny cr}} \mapsto \mathfrak R^{\mbox{\tiny cr}}$ is a decoder of VAE. Here, $\textbf{z}^{\mbox{\tiny cr}} \in \mathbb{R}^{d^{\mbox{\tiny cr}}}$ is a $d^{\mbox{\tiny cr}}$-dimensional latent variable given by $\textbf{z}^{\mbox{\tiny cr}} = \mE(\mathfrak R^{\mbox{\tiny cr}})$ and $\mE^{\mbox{\tiny cr}} : \mathfrak R^{\mbox{\tiny cr}} \mapsto \textbf{z}^{\mbox{\tiny cr}}$ is an encoder of VAE. The maps $(\mE^{\mbox{\tiny cr}},\mD^{\mbox{\tiny cr}})$ and $\Phi^{\mbox{\tiny cr}}$ are trained in the same method presented in \eqref{VAE_mini} and \eqref{nonli_mini} using cranial landmarks $\mathfrak R^{\mbox{\tiny cr}}$. The detailed architectures of $(\mE^{\mbox{\tiny cr}},\mD^{\mbox{\tiny cr}})$ and $\Phi^{\mbox{\tiny cr}}$ are illustrated in Figure \ref{coarse_to_fine_cr}.

\section{Results}
\subsection{Dataset and experimental settings}
Our experiment used a dataset containing 24 paired data (multi-detector CT images and landmark data) and 229 anonymized landmark data. This dataset was provided by Yonsei University, Seoul, Korea. The paired dataset was obtained from normal Korean adult volunteers (9 males and 15 females; 24.22$\pm$2.91 years old) with skeletal class I occlusion and was approved by the local ethics committee of the Dental College Hospital, Yonsei University (IRB number: 2-2009-0026). All informed consents were obtained from each subject. Among 24 paired data, we used 15 data pairs for training (i.e., $N_{p}=15$) and 9 data pairs for testing. The anonymized landmark dataset with 3D landmark coordinates was acquired in an excel format from 229 anonymized subjects with dentofacial deformities and malocclusions (i.e., $N_{l}=229$). Manual landmarking for both dataset was performed by one of the authors (S.-H. Lee) who is an expert in 3D cephalometry with more than 20 years of experience.

Our deep learning method was implemented with Pytorch \cite{Paszke2019} in a computer system with 4 GPUs (GeForce RTX 1080 Ti), two Intel(R) Xeon(R) CPU E5-2630 v4, and 128GB DDR4 RAM. In the training process, the Adam optimizer \cite{Kingma2014} was consistently adopted, which is known as an effective adaptive gradient descent method. In our experiment, all learning parameters (epoch and learning rate) were empirically selected as optimal values via $k$-fold cross validation process \cite{Goodfellow2016}.

\subsection{Experimental results}
\subsubsection{Results of reference landmark detection}
The detection of the 10 reference landmarks ($\mathfrak R_\#$) provided very accurate and robust results (see Table \ref{reference_result} and Figure \ref{final_graph}). These results almost meet clinical requirements, while the intra-observer repeatability is with a precision less than 1 mm and the overall median inter-observer precision is approximately 2 mm in the 3D landmarking system \cite{Pittayapat2016}.

By using reference landmarks, we normalized the landmark data via uniform scaling by fixing the cranial volume of each subject as the average value of the cranial volume for the training dataset.

\begin{table} [t!]
	\centering
	\footnotesize
	\setlength\arrayrulewidth{0.5pt}
	\begin{tabular}{c|c|c|c|c|c}
		Landmark & ANS & Bregma & CFM & Or (L) & Po (L) \\ \hline
		Mean$\pm$SD(mm) & 1.2 $\pm$ 0.45 & 1.9 $\pm$ 0.53 & 2.32 $\pm$ 0.88 & 1.6 $\pm$ 0.335 & 2.21 $\pm$ 0.87 \\ \hline		
		Landmark & Na & Or (R) & Po (R) & ML (L) & ML (R) \\ \hline
		Mean$\pm$SD(mm) & 1.73 $\pm$ 0.565 & 1.3 $\pm$ 0.38 & 1.63 $\pm$ 1.045 & 1.96 $\pm$ 0.65 & 1.72 $\pm$ 0.77 \\
		
	\end{tabular}
	\caption{Detection error evaluation of 10 reference landmarks. Most of the landmarks are annotated almost within clinical requirements.}
	\label{reference_result}
\end{table}

\subsubsection{Results of the initial local-to-global detection}

\begin{figure}[t!]
	\centering
	\includegraphics[width=0.9\textwidth]{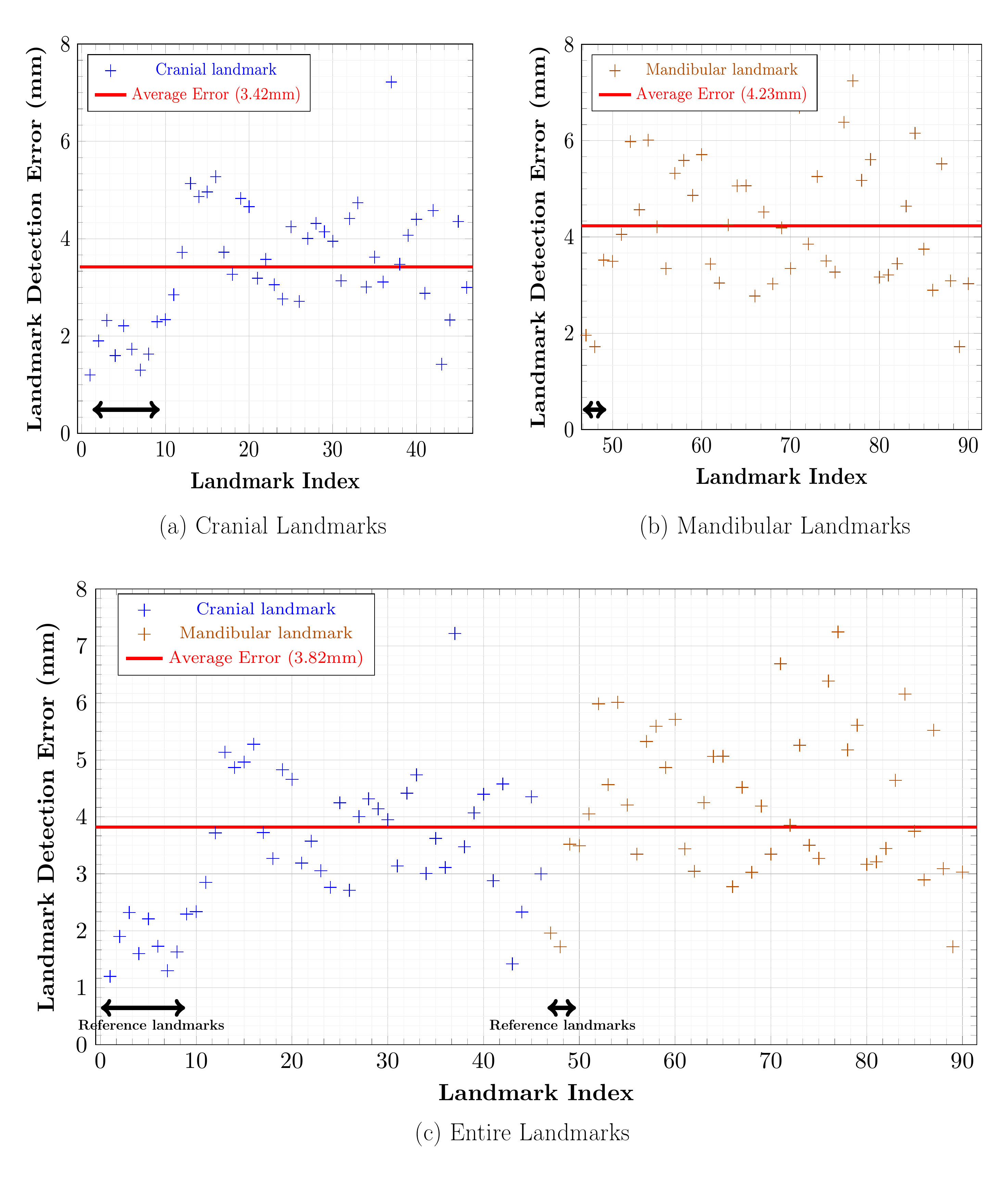}
	\caption{Localization errors (mm) of roughly estimated 90 cephalometric landmarks for the 9 test data using VAE. Blue dots denote the 3D distance error for cranial landmarks, and brown dots represent the 3D distance error for mandibular landmarks. Red line in each diagram represents the average point-to-point error for landmarks included in the diagram.}
	\label{vae_graph}
\end{figure}

\begin{figure}[t!]
	\centering
	\subfigure[Cranial Landmarks]{
		\includegraphics[width=.3\columnwidth]{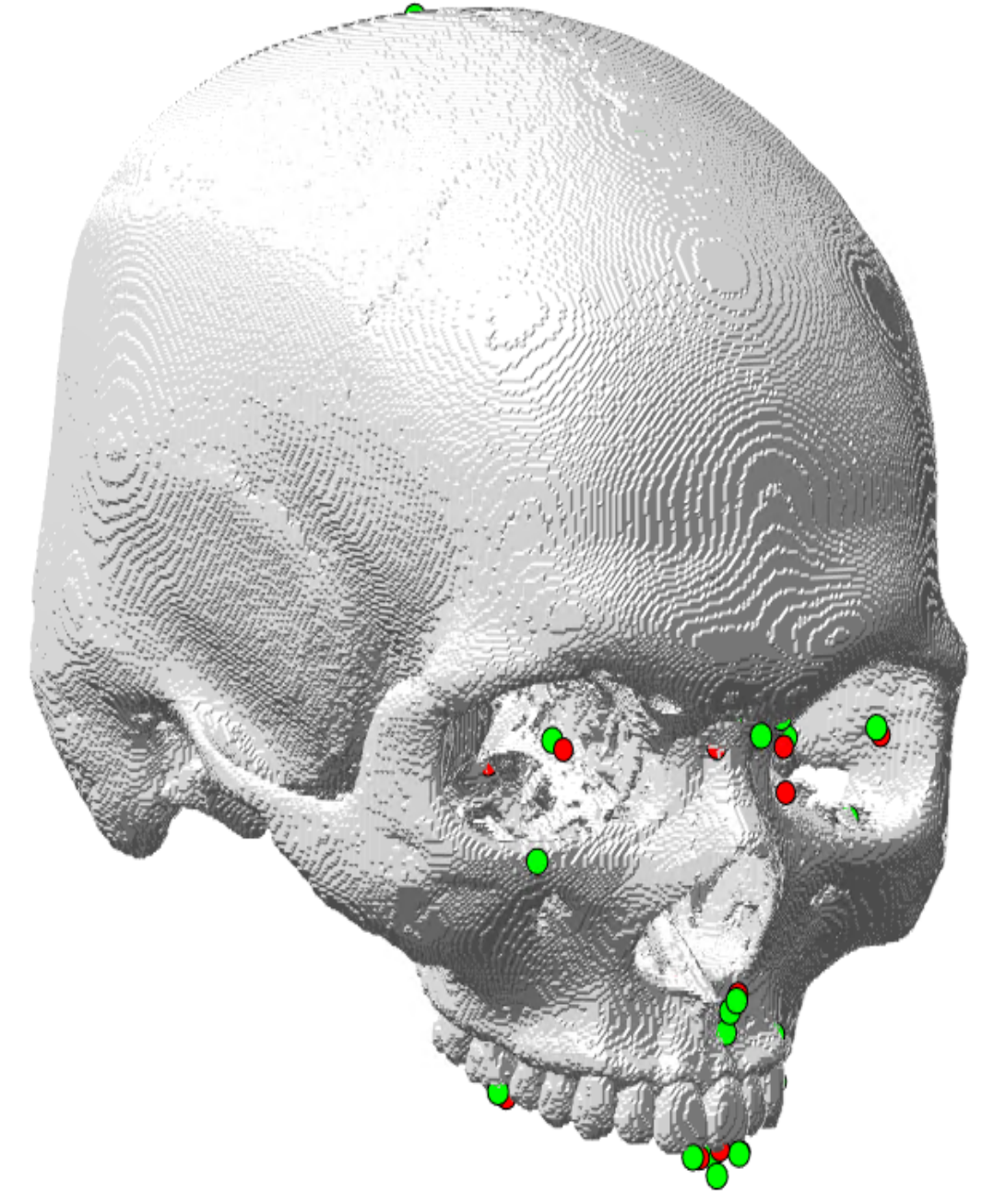}
	}
	~~~~~~~~~~~
	\subfigure[Mandibular Landmarks]{
		\includegraphics[width=.3\columnwidth]{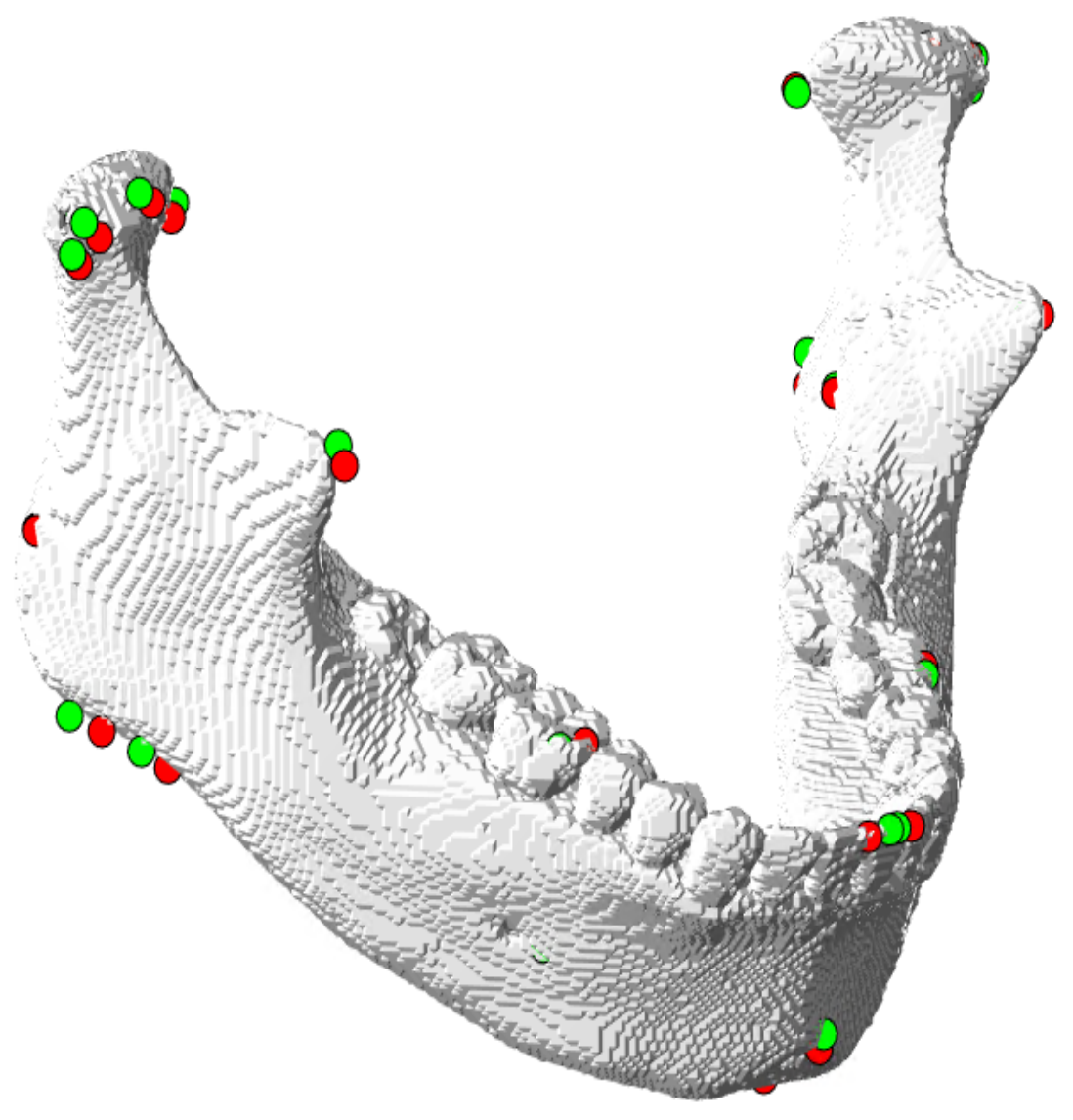}
	}
	\caption{Qualitative evaluation of detection for (a) cranial landmarks and
		(b) mandibular landmarks. The red and green dots denote the ground truth and detected output landmarks respectively.
	}
	\label{fig_final_quali_eval}
\end{figure}

To conduct the initial local-to-global estimation explained in Section \ref{stage2}, 9-dimensional representation was used (i.e., $d=9$ and $\textbf{z} \in \mathbb{R}^{9}$). The VAE $(\mE,\mD)$ was trained using 45000 epochs, a full batch-size, and a $0.001$ learning rate. The nonlinear map $\Phi$ was trained with 11000 epochs, a full batch-size, and a $0.0001$ learning rate.

For each landmark, Figure \ref{vae_graph} shows the performance evaluation achieved using 9 test data with respect to the averaged 3D point-to-point error. The mean detection error was 3.42 mm for the cranial landmarks (Figure  \ref{vae_graph}(a)), 4.23 mm for the mandibular landmarks (Figure \ref{vae_graph}(b)), and 3.82 mm for all landmarks (Figure \ref{vae_graph}(c)). The error of the cranial landmark estimation was much smaller than that of the mandibular landmark estimation.

\subsubsection{Result for coarse-to-fine detection}
\paragraph{Mandibular landmark detection}
\begin{figure}[t!]
	\centering
	\includegraphics[width=.9\textwidth]{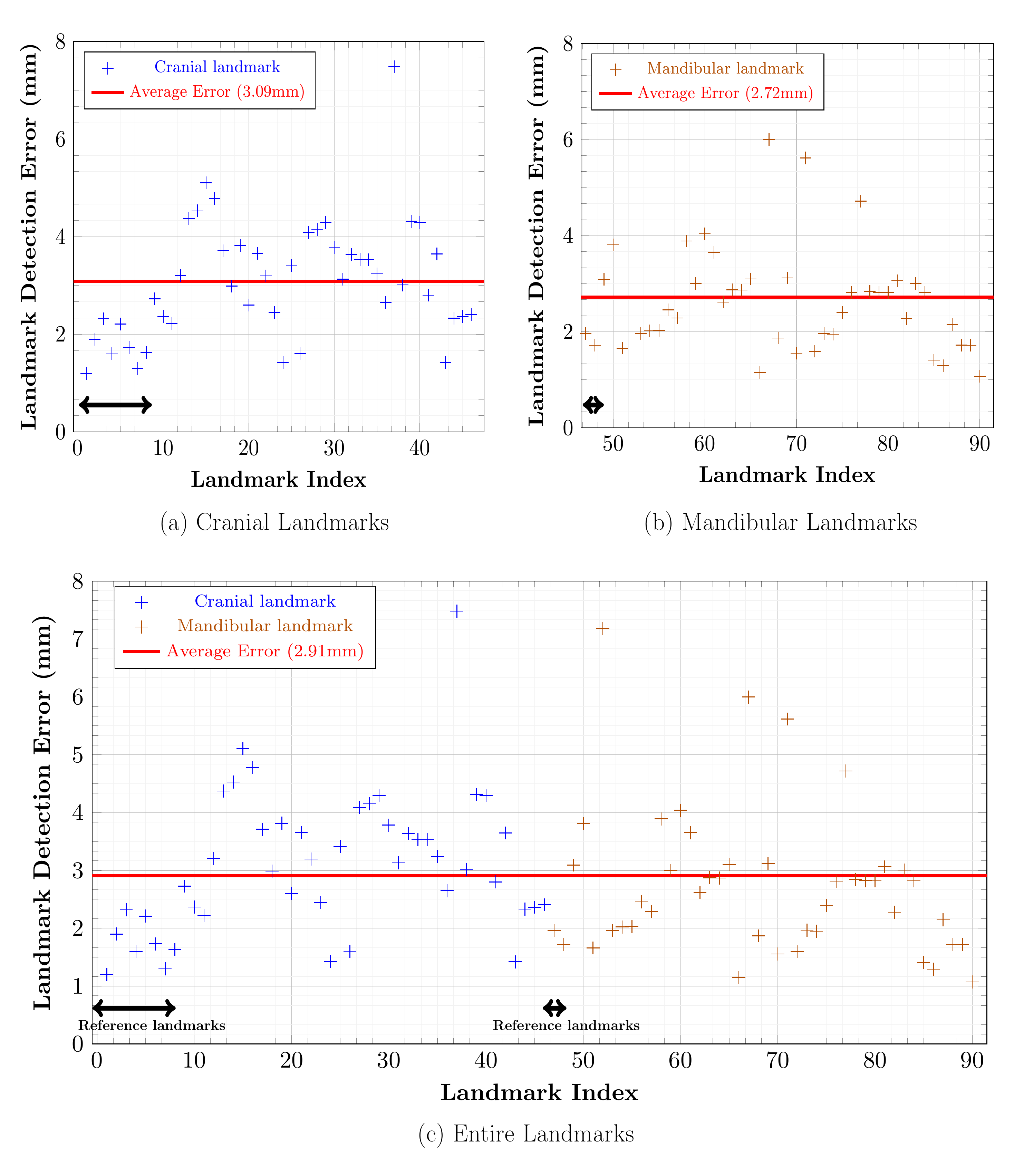}
	\caption{Final localization errors (mm) of 90 cephalometric landmarks for the 9 test data. Blue dots denote the 3D distance error for cranial landmarks, and brown dots represent the 3D distance error for mandibular landmarks. Red line in each diagram represents the average point-to-point error for landmarks included in the diagram.}
	\label{final_graph}
\end{figure}

For fine detection of the mandibular landmarks, 3D image patches were extracted with size of $80\times80\times80$ voxels ($\approx 4 \times 4 \times 4 {\mbox{ cm}}^3$). To generate the training data in \eqref{patch_training_3D}, the center location of patch was varied to cover 2 times the maximum error of the initial estimation of $\mathfrak R_{\diamond}$ for the training data. Using the parameters of 20000 epochs, a full batch size, and a 0.0001 learning rate, nine 3D CNNs were trained.

\begin{figure}[t!]
	\centering
	\includegraphics[width=1\textwidth]{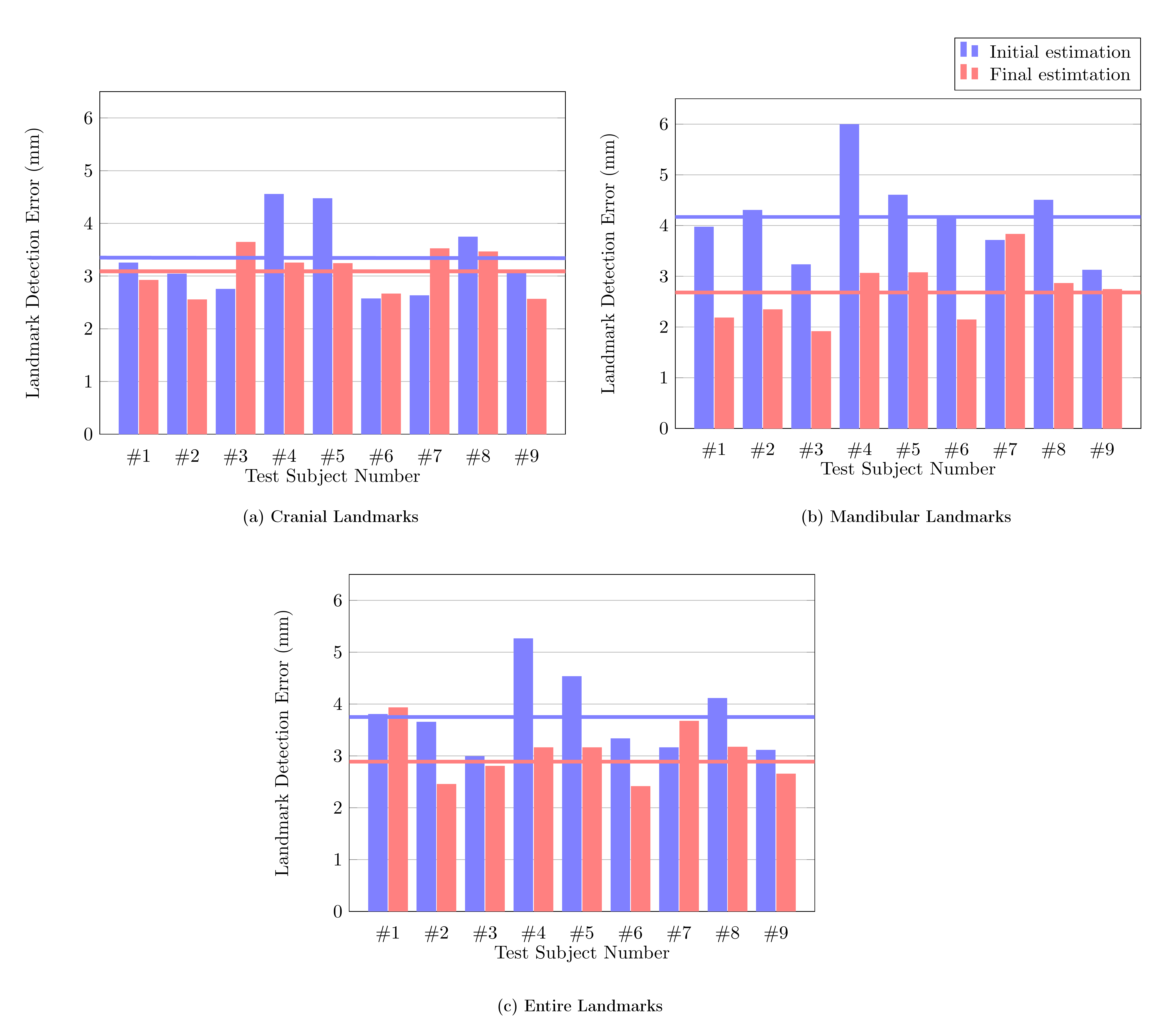}
	\caption{Bar graphs show average point-to-point error of cranial, mandibular and entire landmarks for each of the 9 test data. Blue bars represent the error from the initial estimation, red bars represent the error from the final estimation, and the lines represent average error for all the test data.}
	\label{bar_graph_per_patient}
\end{figure}

Figures \ref{fig_final_quali_eval}(b) and \ref{final_graph}(b) show the qualitative and quantitative results of the 3D CNNs. The mean 3D distance error decreased to 2.72 mm when compared to the initial detection error of 4.23 mm (Figure \ref{vae_graph}(b)). According to results shown in Table \ref{table1}, the proposed method achieved an error range of 1 to 4 mm for the detection of most landmarks. In addition, as shown in Figure \ref{bar_graph_per_patient}(b), the proposed method significantly reduced the mean and variance of error for the test subjects, compared to the initial detection.

\paragraph{Cranial landmark detection}
\begin{figure} [t!]
	\centering
	\includegraphics[width=0.95\textwidth]{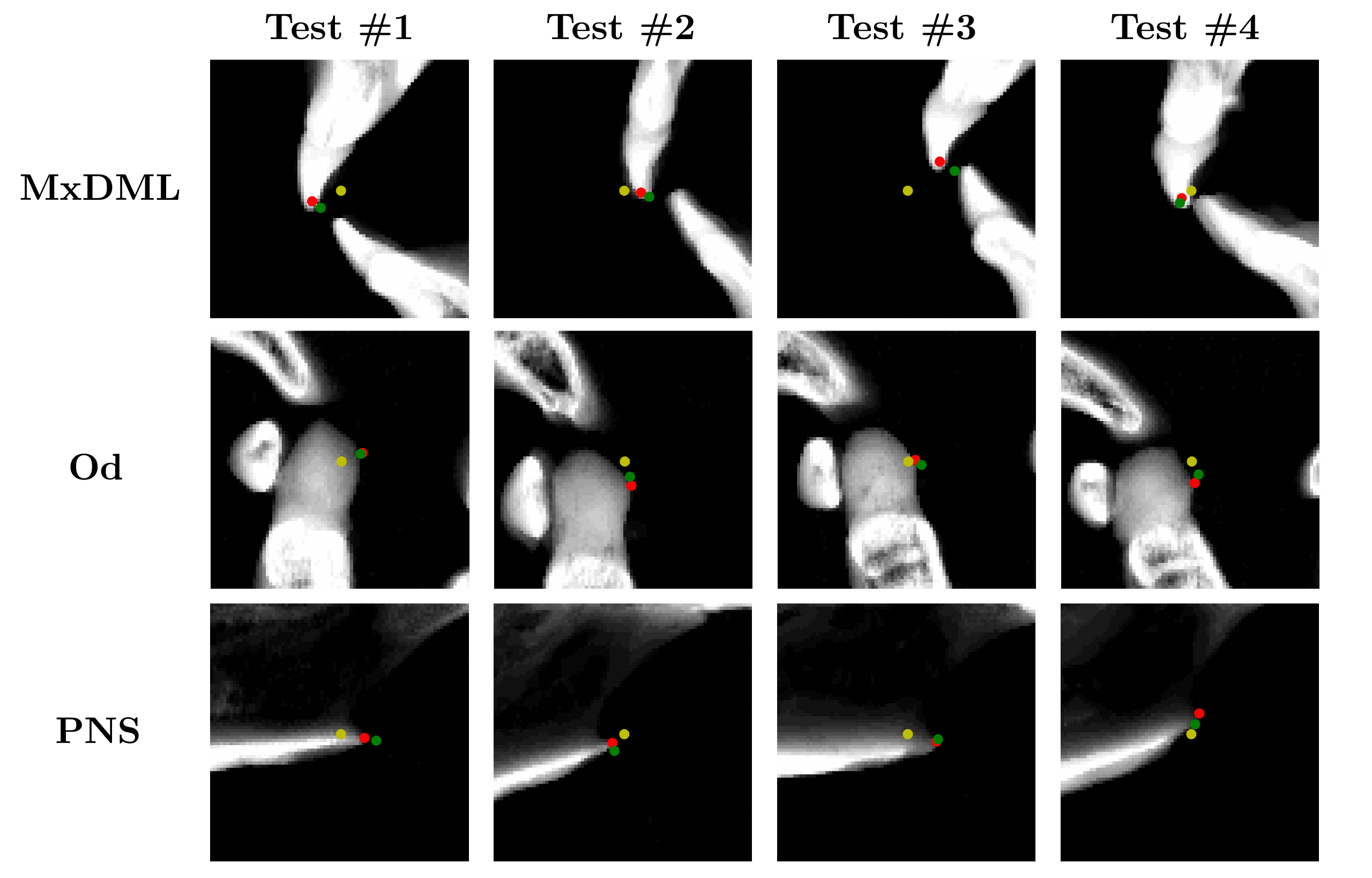}
	\caption{Results of coarse-to-fine landmark detection on 2D patch. Yellow dot is the output of coarsely detected VAE output. Green dot is the output of detection using patch-based CNN. Red dot is the ground truth. }
	\label{patch_result}
\end{figure}

\begin{table} [t!]
	\centering
	\small
	\setlength\arrayrulewidth{0.5pt}
	\begin{tabular}{c||c|c|c}
		Landmark name & Initial error & 2D CNN error & Difference (mm) \\ \hline
		MxDML & 2.76 & 1.42 & -1.34 \\ \hline
		Od & 4.24 & 3.41 & -0.83 \\ \hline
		PNS & 2.71 & 1.60 & -1.11
	\end{tabular}
	\normalsize
	\caption{Error evaluation of the landmarks on the midsagittal plane. Initial error and 2D CNN error are presented in the table. It shows that the errors are reduced after applying 2D CNN.}
	\label{2d_cnn_quanti_eval}
\end{table}

\begin{table}[t!]
	\centering
	\subtable[Initial Estimation]{
		\setlength\arrayrulewidth{0.5pt}
		\begin{tabular}{c||c|c|c|c|c|c|c}
			Error (mm) & 1 - 2 & 2 - 3 & 3 - 4 & 4 - 5 & 5 - 6 & 6 - $\space$ & Total \\ \hline
			Cranium &  7 & 10 & 12 & 14 & 2 & 1 & 46 \\ \hline
			Mandible & 3 & 2 & 16 & 8 & 10 & 5 & 44 \\ \hline
			Total & 10 & 12 & 28  & 22  & 12  & 6 & 90
		\end{tabular}
	}
	\centering
	\subtable[Final Estimation]{
		\setlength\arrayrulewidth{0.5pt}
		\begin{tabular}{c||c|c|c|c|c|c|c}
			Error (mm) & 1 - 2 & 2 - 3 & 3 - 4 & 4 - 5 & 5 - 6 & 6 - $\space$ & Total \\ \hline
			Cranium & 9 & 13 & 14 & 8 & 1 & 1  & 46 \\ \hline
			Mandible &  15 & 15 & 9 & 2  & 2  & 1  & 44 \\ \hline
			Total & 24 & 28  & 23  & 10 & 3  & 2  & 90
		\end{tabular}
	}
	\caption{Error tables show the number of landmarks on cranium and mandible that belong to each error range for initial estimation and final estimation.}
	\label{table1}
\end{table}

To generate the partially integrated image $\x^{\mbox{\tiny mid}}$, we set the interval for the truncated volume as $\pm$ 7.5 mm $v_1$-directionally from the midsagittal plane. Next, 2D image patches were cropped into sizes of $80\times80$ pixels ($\approx 4 \times 4 {\mbox{ cm}}^2$). For training the 2D CNNs, we used the learning parameters of 20000 epochs, a full batch-size, and a 0.0001 learning rate.

In Figure \ref{patch_result} and Table \ref{2d_cnn_quanti_eval}, qualitative and quantitative evaluations of the 2D CNN-based detection of three cranial landmarks on the midsagittal plane are provided. The detection achieved relatively accurate annotation on the three target landmarks.

For the estimation of all cranial landmarks, VAE $(\mE^{\mbox{\tiny cr}},\mD^{\mbox{\tiny cr}})$ was trained with 80000 epochs, a full batch, and a $0.001$ learning rate. The map $\Phi^{\mbox{\tiny cr}}$ was trained with 17000 epochs, a full batch-size, and a $0.0001$ learning rate. The latent dimension was set as 15 (i.e., $d^{\mbox{\tiny cr}}=15$, $z^{\mbox{\tiny cr}} \in \mathbb{R}^{15}$).

Figures \ref{fig_final_quali_eval}(a) and \ref{final_graph}(a) show the final cranial landmark estimation results in qualitative and quantitative formats. The mean detection error for all cranial landmarks was 3.09 mm, which decreased from the initial estimation error of 3.68 mm (Figure \ref{vae_graph}(a)). As shown in Table \ref{table1}, the error for most cranial landmarks fell within the range of 1 to 4 mm.

Considering all landmarks, our proposed method achieved an error of 2.91 mm (Figure \ref{final_graph}(c)), which is much lower than the initial detection error of 3.82 mm (Figure \ref{vae_graph}(c)).

\section{Discussion}
The proposed method employed coarse-to-fine detection, where appropriate strategies for mandibular and cranial landmarks were considered for their different properties. The experiments confirmed the good performance of the proposed method, even when the size of the training dataset is limited. As the amount of training data increases, we expect the detection accuracy to be further improved.

The human skull morphology follows certain patterns and the positions of landmarks are closely interrelated. To learn a low dimensional representation that is strongly associated with the factors determining skull morphology, the proposed method adopted VAE. In our empirical experiment, as shown in Figures \ref{vae_graph}(a) and \ref{vae_graph}(b), the VAE-based approach provided more accurate results for the cranial landmark detection due to the rigid property of the cranium compared to the mandible, which has large shape variance. Among the cranial landmarks, the positional estimation of the SC obtained from the relation learned via VAE exhibited the lowest accuracy (see Figure \ref{final_graph}). This appears to have occurred because the summit position of the cranium (SC) may weakly depend on the positions of other landmarks. A rigorous factor analysis using VAE may be provided in future research.

Recently, as concerns about the radiation doses have increased, there have been attempts to use dental cone-beam CT for cephalometric analysis instead of the conventional multi-detector CT because cone-beam CT utilizes a much lower radiation dose than multi-detector CT. The investigation of an automated 3D landmarking system for cone-beam CT will therefore be a topic of our future research.

\section{Conclusion}
This paper proposes a fully automatic landmarking system for 3D cephalometry in 3D CT. The proposed method provides the accurate and reliable identification of cephalometric landmarks that can be used in subsequent clinical studies, such as in the development of morphometrical guidelines for diagnosis, surgical planning, and the treatment of craniofacial diseases. The proposed semi-supervised method is designed to use many anonymized landmark dataset to address the severe shortage of training CT data. Currently, only 24 CT data pairs are available due to legal and ethical restrictions on medical data, while approximately 200 anonymized landmark data are available.

The proposed method has the potential to alleviate experts' hectic workflow by introducing an automated cephalometric landmarking with high accuracy.
In clinical practice, our method allows all 3D landmarks to be estimated from partial information obtained via 3D CT data. Although the error level of some landmarks does not meet the requirement of clinical applications (less than 2 mm), the proposed method may still aid in decisions of clinicians in determining landmark positions, thereby improving their working processes.

\section*{Acknowledgements}
This research was supported by a grant of the Korea Health Technology R$\&$D Project through the Korea Health Industry Development Institute (KHIDI), funded by the Ministry of Health $\&$ Welfare, Republic of Korea (grant number : HI20C0127).

\bibliographystyle{wileynum}
\bibliography{\jobname}

\appendix
\section{About 90 cephaometric landmarks} \label{appenA}
\centering
\scriptsize
\begin{longtable}[t!]{c|c|c|c} \hline
	Location&Index&Landmark& Description \\ \hline
	cranium & 1 & ANS & Anterior nasal spine (reference) \\ \hline
	cranium & 2 & Bregma & Bregma  (reference) \\ \hline
	cranium & 3 & CFM & Center of foramen magnum (reference) \\ \hline
	cranium & 4 & Or (L) & Left orbitale (reference) \\ \hline
	cranium & 5 & Po (L) & Left porion (reference) \\ \hline
	cranium & 6 & Na & Nasion (reference) \\ \hline
	cranium & 7 & Or (R) & Right orbitale (reference) \\ \hline
	cranium & 8 & Po (R) & Right porion (reference) \\ \hline
	cranium & 9 & $\sharp$16 tip & The mesiobuccal cusp tip of maxillary right first molar \\ \hline
	cranium & 10 & $\sharp$26 tip & The mesiobuccal cusp tip of maxillary left first molar \\ \hline
	cranium & 11 & ANS' & Constructed ANS point \\ \hline
	cranium & 12 & AO & Anterior occlusal point \\ \hline
	cranium & 13 & FC & Falx cerebri \\ \hline
	cranium & 14 & Clp (L) & Left posterior clinoid process \\ \hline	
	cranium & 15 & EC (L) & Left eyeball center \\ \hline
	cranium & 16 & FM (L) & Left frontomaxillary suture \\ \hline
	cranium & 17 & Hyp (L) & Left hypomochlion \\ \hline
	cranium & 18 & M (L) & Left junction of nasofrontal, maxillofrontal, and maxillonasal sutures \\ \hline
	cranium & 19 & NP (L) & Left nasopalatine foramen \\ \hline
	cranium & 20 & Pti (L) & Left inferior pterygoid point \\ \hline
	cranium & 21 & Pts (L) & Left superior pterygoid point \\ \hline
	cranium & 22 & U1 apex (L) & Left upper incisal apex \\ \hline
	cranium & 23 & U1 tip (L) & Left upper incisal tip \\ \hline
	cranium & 24 & MxDML & Maxillary dental midline \\ \hline
	cranium & 25 & Od & Odontoid process \\ \hline
	cranium & 26 & PNS & Posterior nasal spine \\ \hline
	cranium & 27 & Clp (R) & Right posterior clinoid process \\ \hline
	cranium & 28 & EC (R) & Right eyeball center \\ \hline	
	cranium & 29 & FM (R) & Right frontomaxillary suture \\ \hline
	cranium & 30 & Hyp (R) & Right hypomochlion \\ \hline
	cranium & 31 & M (R)  & Right junction of nasofrontal, maxillofrontal, and maxillonasal sutures \\ \hline
	cranium & 32 & Np (R) & Right nasopalatine foramen \\ \hline
	cranium & 33 & Pti (R) & Right inferior pterygoid point \\ \hline
	cranium & 34 & Pts (R) & Right superior pterygoid point \\ \hline
	cranium & 35 & U1 apex (R) & Right upper incisal apex \\ \hline
	cranium & 36 & U1 tip (R) & Right upper incisal tip \\ \hline
	cranium & 37 & SC & Summit of cranium \\ \hline
	cranium & 38 & mid-Clp & \makecell{Midpoint between right and left \\ posterior clinoid point} \\ \hline
	cranium & 39 & mid-EC  & Midpoint between EC (L) and EC (R) \\ \hline
	cranium & 40 & mid-FM  & Midpoint between FM (L) and FM (R) \\ \hline
	cranium & 41 & mid-M  & Midpoint between M (L) and M (R) \\ \hline
	cranium & 42 & mid-Np  & Midpoint between Np (L) and Np (R) \\ \hline		
	cranium & 43 & mid-Or  & Midpoint between Or (L) and Or (R) \\ \hline
	cranium & 44 & mid-Po  & Midpoint between Po (L) and Po (R) \\ \hline
	cranium & 45 & mid-Pti  & Midpoint between Pti (L) and Pti (R) \\ \hline
	cranium & 46 & mid-U1 tip  & Midpoint between U1 tip (L) and U1 tip (R) \\ \hline
	mandible & 47 & MF (L) & Left mental foramen (reference) \\ \hline
	mandible & 48 & MF (R) & Right mental foramen (reference) \\ \hline
	mandible & 49 & $\sharp$36 tip & The mesiobuccal cusp tip of mandibular left first molar \\ \hline
	mandible & 50 & $\sharp$46 tip & The mesiobuccal cusp tip of mandibular right first molar \\ \hline
	mandible & 51 & CON (L) & Left condylar point \\ \hline
	mandible & 52 & COR (L) & Left coronoid point \\ \hline
	mandible & 53 & Cp (L) & Left posterior condylar point \\ \hline
	mandible & 54 & Ct-in (L) & Left medial temporal condylar point \\ \hline
	mandible & 55 & Ct-mid (L) & Midpoint between left Ct-in and Ct-out \\ \hline
	mandible & 56 & Ct-out (L) & Left lateral temporal condylar point \\ \hline
	mandible & 57 & F (L) & Left mandibular foramen \\ \hline
	mandible & 58 & Go-in (L) & Left inferior gonion point \\ \hline
	mandible & 59 & Go-mid (L) & Midpoint between left posterior and inferior gonion point \\ \hline
	mandible & 60 & Go-post (L) & Left posterior gonion point \\ \hline	
	mandible & 61 & L1 apex (L) & Root apex of left mandibular central incisor \\ \hline
	mandible & 62 & L1 tip (L) & Incial tip midpoint of left mandibular central incisor \\ \hline
	mandible & 63 & LCP (L) & Left lateral condylar point \\ \hline
	mandible & 64 & MCP (L) & Left medial condylar point \\ \hline
	mandible & 65 & a-Go notch (L) & Left antegonial notch \\ \hline
	mandible & 66 & mid-F MF (L) & Midpoint between left mandibular foramen and mental foramen \\ \hline
	mandible & 67 & Me (anat) & Anatomical menton \\ \hline
	mandible & 68 & MnDML & Mandibular dental midline \\ \hline
	mandible & 69 & Pog & Pogonion \\ \hline
	mandible & 70 & CON (R) & Right condylar point \\ \hline
	mandible & 71 & COR (R) & Right coronoid point \\ \hline
	mandible & 72 & Cp (R) & Right posterior condylar point \\ \hline
	mandible & 73 & Ct-in (R) & Right medial temporal condylar point \\ \hline
	mandible & 74 & Ct-mid (R) & Midpoint between right Ct-in and Ct-out \\ \hline
	mandible & 75 & Ct-out (R) & Right lateral temporal condylar point \\ \hline	
	mandible & 76 & F (R) & Right mandibular foramen \\ \hline
	mandible & 77 & Go-in (R) & Right inferior gonion point\\ \hline
	mandible & 78 & Go-mid (R) & Midpoint between right posterior and inferior gonion point \\ \hline
	mandible & 79 & Go-post (R) & Right posterior gonion point \\ \hline
	mandible & 80 & L1 apex (R) & Root apex of right mandibular central incisor \\ \hline
	mandible & 81 & L1 tip (R) & Incial tip midpoint of right mandibular central incisor \\ \hline
	mandible & 82 & LCP (R) & Right lateral condylar point \\ \hline
	mandible & 83 & MCP (R) & Right medial condylar point \\ \hline
	mandible & 84 & a-Go notch (R) & Right antegonial notch \\ \hline
	mandible & 85 & mid-F MF (R) & Midpoint between right mandibular foramen and mental foramen \\ \hline
	mandible & 86 & mid-Cp & Midpoint between right and left posterior condylar point \\ \hline
	mandible & 87 & mid-F & Midpoint between F (L) and F (R) \\ \hline
	mandible & 88 & mid-L1 tip & Midpoint between L1 tip (L) and L1 tip (R) \\ \hline		
	mandible & 89 & mid-MF & Midpoint between MF (L) and MF (R) \\ \hline
	mandible & 90 & \makecell{midpoint of \\ mid-F MF (R/L)} & Midpoint between mid-F MF (R) and mid-F MF (L) \\ \hline
	\caption{Entire 90 cephalometric landmarks and their index.}
	\label{90landmark}
\end{longtable}

\end{document}